\documentclass[conference]{IEEEtran}
\IEEEoverridecommandlockouts

\usepackage[ruled,linesnumbered,noend]{algorithm2e}
\usepackage{balance}  
\usepackage{amsmath}
\usepackage{mathptmx}
\usepackage{url}
\usepackage{color}
\usepackage{graphicx}
\usepackage{microtype} 
\usepackage{subfigure}
\usepackage{tabularx} 	
\usepackage{hhline}		
\usepackage{multirow}

\graphicspath{{figures/}}

\newcommand{\IN}{$\in$ }

\newcommand{\DATASET}{$D$}
\newcommand{\FEATURE}{$F$}
\newcommand{\feature}{$f^x$}
\newcommand{\nfeature}{$|F|$}
\newcommand{\INSTANCE}{$I$}

\newcommand{\instance}{$i_y$}
\newcommand{\pinstance}{$i'_y$}
\newcommand{\ninstance}{$|I|$}
\newcommand{\XXVALUE}{$V'^x$}
\newcommand{\XVALUE}{$V^x$}
\newcommand{\xvalue}{$v^x_z$}
\newcommand{\nxvalue}{$|V^x|$}
\newcommand{\XLABEL}{$L$}
\newcommand{\xlabel}{$l$}

\newcommand{\FVS}{Value Selection}
\newcommand{\Fvs}{Value selection}
\newcommand{\fvs}{value selection}
\newcommand{\fvsabbrv}{VS}
\newcommand{\methodA}{PVS}
\newcommand{\methodB}{P$^{+}$VS}
\newcommand{\methodAs}{Probabilistic}

\newcommand{\Ndataset}{10}
\newcommand{\NBaseline}{nine}

\newcommand{\bigVspace}{\vspace{-0.2in}}
\newcommand{\smallVspace}{\vspace{-0.1in}}

\def\plaintitle{Probabilistic \FVS{} for Space Efficient Model}

\usepackage{amssymb}
\usepackage{lineno}

\usepackage[figuresright]{rotating}

\usepackage{amsmath,amsthm,amssymb,scrextend}
\usepackage{fancyhdr}

\renewcommand{\qed}{\hfill$\blacksquare$}

\renewenvironment{proof}{\begin{addmargin}[1em]{0em}\begin{newproof}}{\end{newproof}\end{addmargin}\qed}

\def\BibTeX{{\rm B\kern-.05em{\sc i\kern-.025em b}\kern-.08em
    T\kern-.1667em\lower.7ex\hbox{E}\kern-.125emX}}
\begin{document}

\title{\plaintitle}

\author{\IEEEauthorblockN{Gunarto Sindoro Njoo, Baihua Zheng}
\IEEEauthorblockA{\textit{Living Analytics Research Centre} \\
\textit{Singapore Management University}\\
Singapore \\
gunarton@smu.edu.sg, bhzheng@smu.edu.sg}
\and
\IEEEauthorblockN{Kuo-Wei Hsu}
\IEEEauthorblockA{
kuowei.hsu@gmail.com}
\and
\IEEEauthorblockN{Wen-Chih Peng}
\IEEEauthorblockA{\textit{Department of Computer Science} \\
\textit{National Chiao Tung University}\\
Hsinchu, Taiwan \\
wcpeng@cs.nctu.edu.tw}
}

\maketitle

\begin{abstract}
An alternative to current mainstream preprocessing methods is proposed: \FVS{} (\fvsabbrv). Unlike the existing methods such as feature selection that removes features and instance selection that eliminates instances, \fvs{} eliminates the values (with respect to each feature) in the dataset with two purposes: reducing the model size and preserving its accuracy. Two probabilistic methods based on information theory's metric are proposed: \methodA{} and \methodB{}. Extensive experiments on the benchmark datasets with various sizes are elaborated. Those results are compared with the existing preprocessing methods such as feature selection, feature transformation, and instance selection methods. Experiment results show that \fvs{} can achieve the balance between accuracy and model size reduction.
\end{abstract}

\section{Introduction}
\label{sec:fvs:intro}

Machine Learning is revolutionizing the Mobile App industry. For example, only 2\% of iPhone users have never used Siri and just 4\% of Android users have never used Google Assistant, as reported by Creative Strategies~\cite{VoiceAssistant}. In other words, 97\% of mobile users are using AI-powered voice assistants. Apple has launched Siri SDK and Core ML and now all developers can incorporate this feature into their apps. Similarly, Google has launched TensorFlow for Mobile. Many signals from the major mobile device manufacturers have also confirmed this~\cite{ML-Mobile}. Lenovo is working on its new smartphone that also performs without an internet connection and executes indoor geolocation and augmented reality; many mobile chip makers (including Apple, Huawei, Qualcomm, and Samsung) are working on hardware dedicated to accelerating machine learning on mobile devices.

There are many desirable advantages of enabling mobile devices to perform machine learning tasks without connecting to the servers, including but not limited to increased security and privacy, no internet connection required, lower latency and so on. However, many machine learning tasks, originally designed to run in computers, require a lot of computational power; and some model files could be huge, which incur an expensive space overhead. On the other hand, mobile devices have significantly less powerful computation capability and smaller storage space, as compared with servers or computers. How to adapt the machine learning algorithms/models to resource-limited mobile devices is a challenging issue. 

In this paper, we focus on \emph{classification}, one of the most common machine learning tasks. To be more specific, we study compact classification methods such as Decision Tree~\cite{Quinlan:1993:CPM:152181}, rule-based method~\cite{Cohen:1995:FER:3091622.3091637}, and Naive Bayesian~\cite{hand2001idiot} but not complex classification models such as support vector machine (SVM)~\cite{Platt1998} or neural network~\cite{Rumelhart:1986:LIR:104279.104293}. This is because storage spaces and computational power are still considered as limited resources in many low-end smartphones. In addition, compact classification methods are able 
to achieve certain accuracy and are easy to implement/interpret. For example, decision trees have been extensively used for bank loan approvals owing to their extreme transparency of rule-based decision-making. In summary, we dedicate this paper to the study of ways that could reduce the complexity of compact classification methods to cut down the storage overhead without losing much accuracy.

Reducing the storage overhead is a common requirement of many tasks. For example, in data mining, \emph{dimension reduction} and \emph{instance selection} are two preprocessing methods that are generally used to reduce the complexity of the data. The former is to reduce the dimensional complexity of the data/model by removing unproductive features/dimensions via \emph{feature selection} or projecting the original feature space into a different feature space with lower dimensionality via \emph{feature transformation}. The latter is to reduce the number of training instances to speed-up the training process, under the assumption that the whole dataset can be represented by a number of core instances without suffering much information loss. 

Motivated by the strengths of both feature selection and instance selection, we explore a new preprocessing method called ``\fvs'' in this paper. For better understanding, let's assume the data are recorded by a two-dimensional table, with the columns referring to different features and each row (or a record) corresponding to the detailed feature values of one instance. Removing a feature is equivalent to delete a complete column from the table, and removing an instance is equivalent to delete a complete row from the table. Instead of removing a complete row or a complete column, \fvs{} adopts a finer granularity, i.e., the intersection between them, which is the values. It evaluates the importance of individual values, but not that of complete rows/columns when selecting data for deletion. We hypothesize that values have different importance in the training process. \FVS{} enables the measurement of importance at the finest granularity (i.e., at the value level) and can remove those values that do not improve the classification accuracy. Consequently, it is expected to be able to achieve a better trade-off between the model size and accuracy. 
%

\begin{figure}
      \centering
      \includegraphics[width=.9\linewidth]{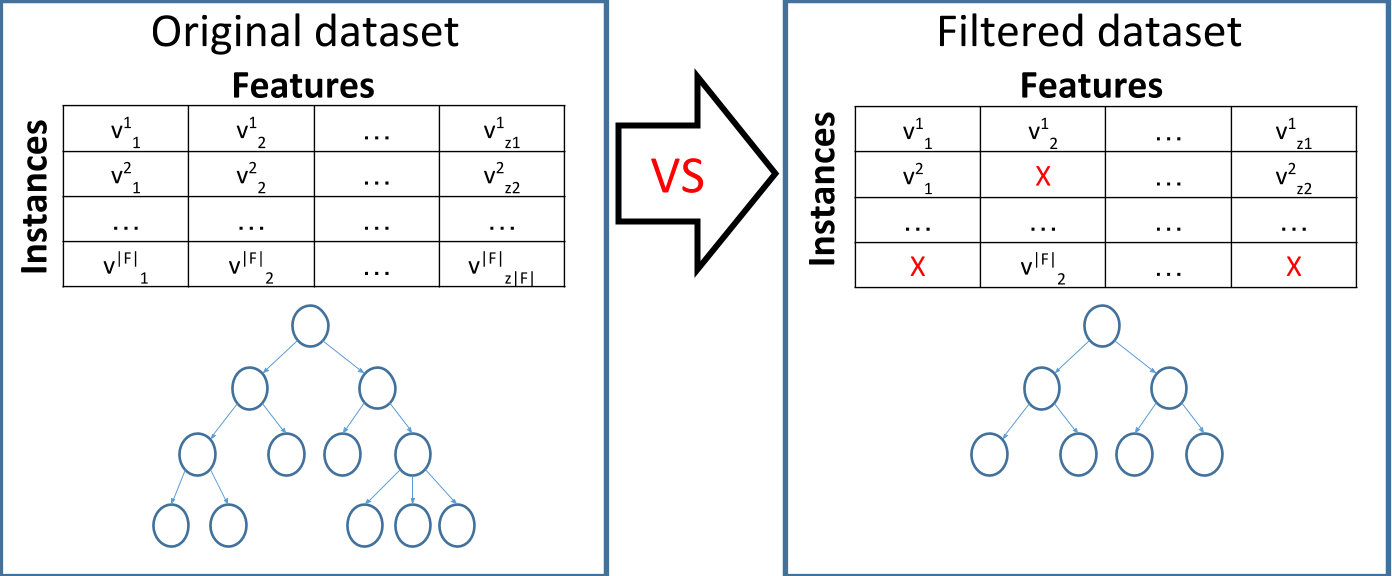}
      \caption{Illustration of the \FVS{} and how it reduces the model size. Left: original data representation. Right: filtered data representation.}
      \label{fig:fvs:illustration}
      \bigVspace
\end{figure}

Here, for illustration purpose, \fvs{} is applied to the Decision Tree algorithm~\cite{Quinlan:1993:CPM:152181} since the rules built in the Decision Tree model are conjunctions of values, which makes \fvs{} relevant to improve the model's performance and to reduce the model's complexity. Moreover, Decision Tree is one of the predictive models that is often used in data mining to solve both the classification task and the regression task~\cite{breiman1984classification}. 
Without losing the generality, \fvs{} can also be implemented on other classification methods that share similar properties with Decision Tree, such as rule-based classifier~\cite{Cohen:1995:FER:3091622.3091637}. In conclusion, Figure~\ref{fig:fvs:illustration} shows the concept of the proposed \FVS{} and illustrates how the model size could be reduced.

The contributions of this work is summarized as follows.
\begin{itemize}
    \item A new preprocessing method called \emph{\FVS{} (\fvsabbrv)} is proposed. Its primary goal is to reduce the model's size without sacrificing the accuracy of the classification model.
    \item Two \fvs{} methods, namely \emph{\methodA{}} and \emph{\methodB{}}, are formulated based on the information theory's metrics. 
    \item A comprehensive experimental study has been performed to compare the proposed methods with \NBaseline{} baselines using \Ndataset{} benchmark datasets~\cite{Dua:2017, Krizhevsky09}. The experimental results show that our methods can reduce the model size substantially and maintain acceptable accuracy.
\end{itemize}

\section{Related Works}
\label{sec:fvs:related}

In this section, we review existing works related to \emph{instance selection} in Section \ref{sec:fvs:related:is} and those related to \emph{dimension reduction} in Section \ref{sec:fvs:related:fs} respectively. 

\subsection{Instance Selection}
\label{sec:fvs:related:is}

Instance selection is often elaborated to minimize the number of training instances when the training computation cost is high, especially with the usage of computationally expensive classifiers such as nearest neighbor and SVM. 
Prior research on the instance selection mostly focuses on the classifier-specific instance selection, while the latest instance selection methods deal with more advanced topics (e.g., unsupervised learning, online learning, and active learning).

\noindent
\textbf{Classifier-specific instance selection.}
Often, instance selection methods are tailored toward a specific classification model that requires high computational costs such as nearest neighbor, SVM, and neural network. Whereas, despite their high computational costs, those classifiers often produce satisfying results in terms of the prediction power.
For the nearest neighbor method, the authors in~\cite{Chou:2006:GCN:1170748.1172164} proposed a generalized CNN (Condensed Nearest Neighbor) method to shrink the training set. The generalization of CNN is performed by adding different absorption criteria: using a threshold multiplied by the minimum norm between arbitrary points. For SVM, the authors in~\cite{akinyelu2017improved} proposed two instance selection methods based on the firefly algorithm~\cite{Yang:2008:NMA:1628847,2013arXiv1308.3898Y} and edge detection method in image processing, respectively. Another work related to SVM \cite{Liu201758} used a geometry-based approach to perform instance selection on SVM. It assumes a spherical class distribution and distributes a decision plane between spheres. Accordingly, it removes the vector points that are not distributed in the adjacent of the two hemispheres, as they are non-support vectors.
For the neural network, the authors in \cite{7727625} proposed an entropy-based method to perform an unsupervised instance selection.  They use entropy to evaluate the information contained in each instance, i.e., instances with higher entropy tend to have more information compared to others.

\noindent
\textbf{General-purpose instance selection.}
Various general purpose instance selection methods have been proposed in the past, including randomized approach, clustering the instances, etc. 
The authors in~\cite{Vitter:1985:RSR:3147.3165} reported a fast randomized approach for instance selection, by using the reservoir concept. A reservoir with a predefined size is used to store the instances. The first batch of data is stored immediately on the reservoir and others are processed using a probabilistic approach. However, this method makes a strong assumption, that is each instance contains the same amount of information, and the more important instances occupy a larger part of the dataset.
The authors in~\cite{dai2011instance} presented three instance selection methods based on the reverse nearest neighbor (RNN) concept.
The authors in~\cite{HAMIDZADEH20151878} proposed an instance selection method using hyper-rectangle clustering~\cite{Salzberg:1991:NHL:104735.104741}. Hyper-rectangles are generated for each class, and the mean of its interior instances is used as the representative instance. 
Recently, the authors in~\cite{ARNAIZGONZALEZ201683} proposed an instance selection with linear time complexity for larger dataset based on the concept of locality-sensitive hashing~\cite{Pauleve:2010:LSH:1814586.1814748} to quickly identify the similarity between instances.

\noindent
\textbf{Instance selection for specific topics.}
Most recent efforts on the instance selection methods are more related to specific problems related to unsupervised learning, online learning, and active learning.
The authors in~\cite{Patra2015} proposed a data structure called \textit{data sphere} to summarize the data and to speed up hierarchical clustering methods. The proposed method, namely \emph{summarized single-link method}, extends the previous work called \emph{single-link}~\cite{sneath1973numerical} to scale it for larger datasets and outperforms the original \emph{single-link} by two orders of magnitude.
The authors in~\cite{Chu:2011:UOA:2020408.2020444} studied the problem of online active learning, by using a binary classification to perform selective labeling in the data streams.

\noindent
\textbf{Discussion.} 
The difference between existing instance selection methods and the proposed \fvs{} methods lies in their goals: instance selection methods reduce the number of the training instances to ease the training process, whereas \fvs{} methods reduce the classification model size. Undoubtedly, a \fvs{} method can also be generalized as an instance selection method if all the values in an instance are removed.
\subsection{Dimension Reduction}
\label{sec:fvs:related:fs}

As stated in Section~\ref{sec:fvs:intro}, we can perform either feature selection or feature transformation to reduce the dimensionality of the feature space. 

\noindent
\textbf{Feature selection}. Selecting relevant features could help classification models to be more accurate and concise. Feature selection methods remove unnecessary features by using various selection criteria. In general, feature selection can be categorized into filter and wrapper approaches based on the inclusion of the classification model in the process. 
In the light of filter approaches, the authors in \cite{Liu1996} built a probabilistic model according to the inconsistency criterion to eliminate the features.
Differently, correlation feature selection (CFS) \cite{Hall1998} performs feature selection on the dataset by evaluating the correlation between features and observing the predictive power of each feature.
Using entropy in mind, the authors in \cite{Largeron:2011:EBF:1982185.1982389} devised a feature selection method on the text data. A term distribution on different categories determines the discriminative power of the term; a term that appears in multiple categories is less discriminative than a term that appears only in one category.
In the light of natural computing, the authors in \cite{PSOSearch2007} utilize particle swarm optimization which is often applied in a continuous search space problem to find the optimal feature subset.
Wrapper feature selection is often used to increase the performance of feature selection, even though it requires longer running time (e.g., \cite{IWSS2009, MOEA2017}). The authors in \cite{MOEA2017} presented a wrapper approach with two objective functions: minimizing the number of selected features and minimizing the root mean squared error (RMSE) of the model learned by Random Forest (RF). A ten-fold cross-validation is used to minimize the risk of overfitting to certain features. Finally, a hybrid approach is presented in \cite{IWSS2009} by adapting an incremental wrapper where the features are ranked first using a filter measure and further evaluated using the wrapper approach. 

\noindent
\textbf{Feature transformation}. Feature transformation could reduce the feature complexity too.  Instead of removing the feature directly as feature selection does, it projects the dataset into a different data space to reduce the dimensional complexity.
Principal component analysis (PCA)~\cite{pearson1901liii} is one of the most famous data analysis tools that could reduce the complexity of the dataset. PCA is calculated by eigenvalue decomposition of a data covariance (or correlation) matrix or singular value decomposition of a data matrix. 
Random projection~\cite{Bingham:2001:RPD:502512.502546} provides an alternate solution to PCA, using a random matrix to transform the original dataset into another data space. The concept is built on top of Johnson-Lindenstrauss lemma~\cite{johnson1984extensions}. That is, the distance between the projected points, which are projected onto a randomly selected subspace, is an approximation of the distance between the points in the original space. In terms of the computation time, random projection is significantly less expensive than PCA, while random projection could yield results comparable to PCA~\cite{Bingham:2001:RPD:502512.502546}.

\noindent
\textbf{Discussion.} 
While most feature selection methods aim to address the curse of dimensionality, they do not consider reducing the model size. Consequently, the classification model built on top of feature selection is only expected to have higher accuracy \cite{DASH1997131}.
On the other hand, feature transformation methods can reduce the model size and maintain acceptable accuracy. However, they produce obscure rules and in general require a long time to compute due to the complex computation.
\section{Preliminaries}
\label{sec:fvs:preliminaries}

In this section, we first define in Section \ref{sec:fvs:pre:terms} the terms and notations that are frequently used in this paper, and then formulate the \fvs{} problem in Section \ref{sec:fvs:pre:problem}.

\subsection{Terms and Notations}
\label{sec:fvs:pre:terms}

\begin{table}[t]
\centering
\footnotesize
\caption{Summary of notations}
\label{tab:fvs:notations}
\begin{tabular}{|l|l|}
\hline
Notation & Meaning \\ \hline
\DATASET & Dataset matrix of \nfeature{} $\times$ \ninstance{}\\
\FEATURE & Feature set with \nfeature indicating the number of features\\
\INSTANCE & Instance set with \ninstance indicating the number of instances\\
\feature & The $x$-th feature / column / dimension \\
\instance & The $y$-th instance / row / record \\
\XVALUE & Collection of all values of $x$-th feature \\
\xvalue & The $z$-th value of $x$-th feature \\
\nxvalue & Number of possible values of $x$-th feature \\
\XXVALUE & Filtered value set of $x$-th feature \\
\XLABEL & Class label set \\
\xlabel & A class label \IN \XLABEL{} \\
$|M|$ & Classification's model size \\
$|M_{o}|$ & Original model's size \\
$|M_{p}|$ & Preprocessed model's size \\
$Acc_o$ & Original model's accuracy \\
$Acc_p$ & Preprocessed model's accuracy \\
$MR$ & Model size reduction \\
$AR$ & Accuracy ratio \\
H(\DATASET) & Entropy of dataset \DATASET \\
H(\DATASET$|$\xvalue) & Conditional entropy of dataset \DATASET{} given value \xvalue{} \\
$IG$(\DATASET, \xvalue) & Information gain in the dataset given the value \xvalue{}\\
$\overline{X}$ & Harmonic mean of accuracy ratio and model size reduction
\\ \hline
\end{tabular}
\end{table}

Table \ref{tab:fvs:notations} summarizes the notations used throughout this paper.
Each instance \instance{} \IN \INSTANCE{} is a list of values and has a class label \xlabel{} \IN \XLABEL{}. It is important to note that an instance might have values corresponding to certain features missing.
Value \xvalue{} is a distinct value that corresponds to a feature \feature{} \IN \FEATURE{}, which might appear once or multiple times in any instance \instance{} \IN \INSTANCE{}. The values \xvalue{}s in each feature \feature{} are non-overlapping and independent to each other. Finally, to simplify the value removal process and to reduce the search space, we discretize all the continuous values into discretized values.

As our main objective is to reduce the model size without losing the accuracy, we adopt the \emph{model size reduction} and the \emph{accuracy ratio} as the major performance metrics. 
%
For the sake of simplicity and consistency, the model size $|M|$ is measured by the total number of leaves/internal nodes throughout the paper. In other words, $|M|$ refers to the total number of rules that are used to represent the model. 
%
%
Without losing the generality, a rule is a path from the root to a leaf in a decision tree. The reduction of the tree size in the model represents the reduction of the model's complexity, which ultimately reduces the size of the classification model. 
Subsequently, model size reduction $MR$ is the normalized difference between the original model size $|M_{o}|$ and the preprocessed model size $|M_{p}|$, as shown in Equation~(\ref{eq:fvs:modelreduction}).
\begin{equation}
MR = \frac{|M_{o}| - |M_{p}|}{|M_{o}|}
\label{eq:fvs:modelreduction}
\end{equation}
The range of model size reduction is $-\infty < MR < 1$ but the typical value range is between 0 and 1. A positive $MR$ value indicates that the preprocessing has successfully reduced the size of the model; a zero $MR$ value indicates that the preprocessing does not reduce the size of the model; and a negative $MR$ value reflects that the preprocessing actually enlarges the model size, which is not desirable. In general, a larger $MR$ value is more preferable. 
%

Similarly, accuracy of a classification model might be changed when the data is preprocessed, whether it is an improvement or a deterioration. 
To evaluate the effectiveness of the preprocessing methods, one could evaluate the difference between the accuracy of the model built on the original data (denoted by $Acc_o$) and the accuracy of the model built on the preprocessed data (denoted by $Acc_p$).
However, the absolute accuracy difference mentioned above cannot reflect the gain or the loss relatively to the original model's accuracy. Thus, a metric called \emph{accuracy ratio ($AR$)} is introduced to quantify the ratio between the preprocessed model's and the original model's accuracy, as explained in Equation~(\ref{eq:fvs:accratio}).
\begin{equation}
AR = \frac{Acc_{p}}{Acc_{o}}
\label{eq:fvs:accratio}
\end{equation}
The range of accuracy ratio is $0 \leq AR < \infty$, in which $AR = 1$ means no accuracy changes, $AR<$1 reflects accuracy deterioration, and $AR>$1 expresses accuracy improvement. Again, a higher $AR$ is more desirable than a lower $AR$.

\subsection{Problem Definition}
\label{sec:fvs:pre:problem}
\begin{figure}
      \centering
      \includegraphics[width=\linewidth]{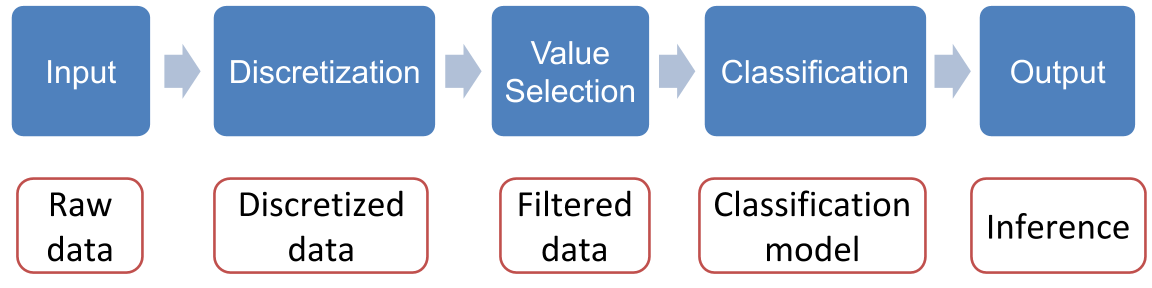}
      \caption{\Fvs{} in the general classification process.} 
      \label{fig:framework}
      \bigVspace
\end{figure}

Figure \ref{fig:framework} explains how the \fvs{} is performed in a data processing pipeline.
The original dataset \DATASET{} needs to be discretized before the \fvs{} stage. Two \fvs{} methods are proposed in this paper: \methodA{} and \methodB{}. Both methods take advantage of the information metric of each value and apply a probabilistic approach based on the information metric's value. Details of both methods are explained in Sections \ref{sec:fvs:method:probabilistic} and \ref{sec:fvs:method:probabilisticplus}, respectively. 
%




  
\section{Methodology}
\label{sec:fvs:method}

In this section, we first introduce an approach to evaluate the important of values for the task of classification; we then present the two proposed value selection approaches, namely \methodA{} and \methodB{}. Finally, we explain the effectiveness proof and intuition behind the proposed solution.

\subsection{The Importance of Values}
\label{sec:fvs:method:fv}

As explained in Section~\ref{sec:fvs:pre:terms}, a value is the intersection of a feature and an instance. For a given feature, it might have values that are relevant to the classification and values that are irrelevant. 
%
%
Thus, simply removing a feature results in losing all the good values that might hurt the classification accuracy.
%
%
Therefore, the proposed \fvs{} methods aim to preserve those good values in each feature. By doing so, \fvs{} ultimately can maintain accuracy and reduce the model size (and hence the model complexity). 
%
%
Naively, to perform \fvs, one can simply remove values randomly.  
However, removal of values does not necessarily induce model size reduction in all cases, as it might introduce the overfitting problem. Thus, we introduce new metrics based on information theory to quantify the importance of each value.

The importance of each value is determined by its predictive power to deduce the class label. However, each value might possess different class distribution, which would complicate the model building process. Accordingly, we propose two types of information metrics, i.e., \emph{entropy} and \emph{information gain}, to discern the predictive power of each value. To ease the explanation, information metric $\iota$ is denoted to be either entropy or information gain for the values' goodness metric. 

Firstly, Shannon entropy \cite{6773024} is adapted to measure the values' goodness; values with lower entropy (i.e., less confusion) tend to be more useful than values that have higher entropy. The adaptation of the Shannon entropy is stated in Equation~(\ref{eq:fvs:entropy}).
\begin{equation}
H(D|v^x_z) = - \sum_{\forall l\in {L}} p^x_{l,z} \log_{|L|} p^x_{l,z}
\label{eq:fvs:entropy}
\end{equation}
Here, probability $p^x_{l,z}$ expresses the probability of class label $l$, given the value \xvalue{} observed corresponding to the feature \feature{}. To be more specific, given a feature \feature{}, there are in total \nxvalue{} different values. For each value \xvalue{} observed in this feature, we can count the probability of this value observed for a given class $l\in {L}$; and the sum of $p^x_{l,z}$ corresponding to different \xvalue{} values in feature \feature{} is one, i.e., $\sum_{v^x_z \in V^x_z} p^x_{i,z} = 1$. The entropy of any value $H(D|v^x_z)$ is in the range of $0$ and $1$. 

Take the sample instance set listed in Table~\ref{tab:samples} as an example. We assume there are in total two classes, i.e., $L = \{0, 1\}$. For feature $f^3$, there are in total three distinct feature values, with $v_1^3 = 2$, $v_2^3 = 1$, and $v_3^3 = -1$ (i.e., $V^3 = \{2, 1, -1\}$). Now, let's derive the entropy of three different values. $H(D|v_1^3) = -p^3_{l=0,1} \log_{2} p^3_{l=0,1} - p^3_{l=1,1} \log_{2} p^3_{l=1,1}$. As there is no instance \instance{} in the class $l=0$ having its feature value in the feature $f^3$ being $2$, $p^3_{l=0,1}=0$ and $p^3_{l=1,1}=1$ because all the instances (i.e., one instance) belong to class $l=1$. Accordingly, $H(D|v_1^3)= -0\log_{2}0-1\log_{2}1=0$. Following the same logic, we have $H(D|v_2^3) = H(D|v_3^3) = -\frac{1}{3}\log_{2}\frac{1}{3}-\frac{2}{3}\log_{2}\frac{2}{3}$ and $H(D|v_3^3) = 0$. In other words, $v_1^3$ and $v_3^3$ are more useful in predicting the classes of instances in feature $f^3$, as compared to $v_2^3$. 

\begin{table}
\centering
\footnotesize
\caption{Sample instances with their corresponding feature values and class labels}
\label{tab:samples}
	\begin{tabular}{|c||c||c|c|c|c|}
	\hline
	instances & class label & $f^1$ & $f^2$  & $f^3$ & $f^4$ \\ \hline
	$i_1$ & $1$ & -     & $1$   & $2$   & $1$ \\ \hline
	$i_2$ & $1$ & $1$   & -     & $1$   & $1$ \\ \hline
	$i_3$ & $0$ & $-1$  & $-2$  & $1$  & $-1$ \\ \hline
	$i_4$ & $0$ & $-1$  & -     & $1$  & $-2$ \\ \hline
	$i_5$ & $1$ & $1$   & $1$   & $-1$  & - \\ \hline
	\end{tabular}
\end{table}


In addition to entropy, information gain~\cite{Quinlan:1986:IDT:637962.637969}, which is widely used in the decision tree, is applied to serve as the other goodness metric for the values. The adaptation of information gain for \fvs{} is presented in Equation~(\ref{eq:fvs:ig}). Note that $H(D) = \sum_{v^x_z \in D} H(D|v^x_z)$. 
\begin{equation}
IG(D,v^x_z) = H(D) - H(D|v^x_z)
\label{eq:fvs:ig}
\end{equation}

In order to ensure the values of information gain are also in the range of $0$ and $1$, we introduce the normalized information gain in Equation~(\ref{eq:fvs:ign}). Different from the entropy, a larger information gain indicates a value with a stronger predictive power and hence is expected to be more important than a value with smaller information gain, for the task of classification.  
%
\begin{equation}
IG_N(D,v^x_z) = \frac{IG(D,v^x_z)}{\displaystyle \max_{\forall v^x_z \in V^x} IG(D,v^x_z)}
\label{eq:fvs:ign}
\end{equation}


\subsection{Probabilistic \FVS}
\label{sec:fvs:method:probabilistic}

The most straightforward way to perform \fvs{} using an information metric $\iota$ (i.e., either the entropy or the information gain) is to discriminate less useful values from the dataset by using a user-defined threshold $\tau$, i.e., removing all the values \xvalue{} with entropy $H(D|v^x_z)>\tau$ or with information gain $IG_N(D,v^x_z)<\tau$. However, determining a threshold for an information metric $\iota$ is not a trivial task~\cite{DBLP:conf/dsaa/NjooPHP14}. 
Therefore, we adopt a probabilistic approach, instead of the threshold-based removal approach, to select the values by using $\iota$ as the probability for a value removal, e.g., values with larger entropy (smaller information gain) are more likely than those with smaller entropy (larger information gain) to be removed. This avoids the threshold selection process and eases the application of \fvs{} in other domains.
In addition, an amplifier hyperparameter $\epsilon$ (with $0 < \epsilon \leq 1$) is used to intensify the value removal probability, where a small $\epsilon$ amplifies \fvs{}'s impact.
In summary, Equation~(\ref{eq:fvs:probabilistic:entropy}) defines the probability of a value \xvalue{} to be removed from the value set \XVALUE{}. It is worth highlighting that Equation~(\ref{eq:fvs:probabilistic:entropy}) unifies the two types of metrics, and a higher probability indicates a higher chance to be removed as the underlying value has either larger entropy or smaller information gain. 
%
%
\begin{equation}
P(V^x \setminus v^x_z) = 
\left\{ \begin{array}{ll}
\frac{H(D|v^x_z)}{\epsilon} & \textrm{if $\iota$ is entropy} \\
\frac{1-IG_{N}(D,v^x_z)}{\epsilon} & \textrm{if $\iota$ is information gain}
\end{array} \right.
\label{eq:fvs:probabilistic:entropy}
\end{equation}
%

\begin{figure*}
      \centering
      \includegraphics[width=.85\textwidth]{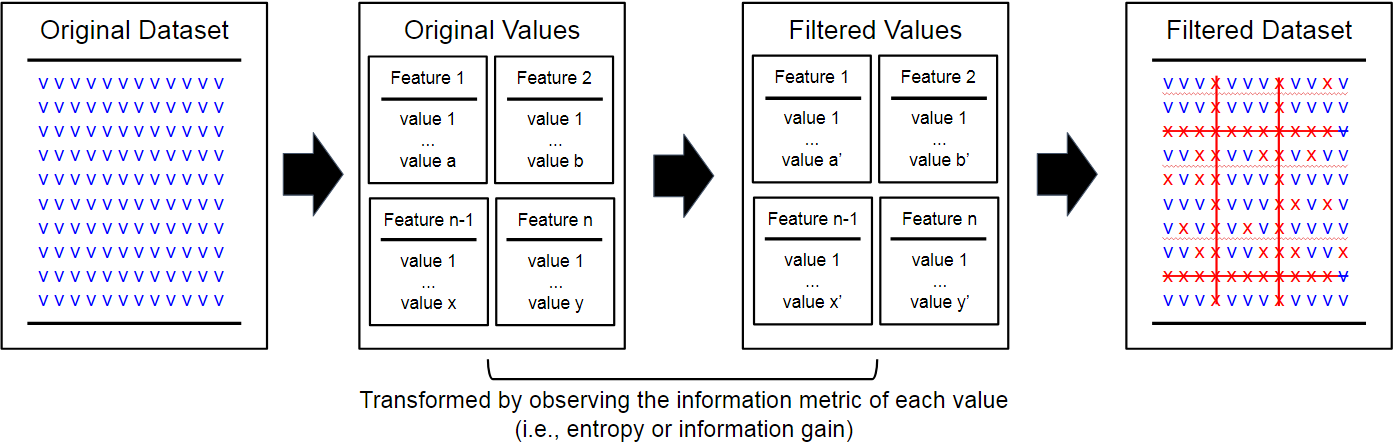}
      \smallVspace
      \caption{Flowchart of the \methodAs{} \FVS{} process. The \textcolor{blue}{blue 'v'} characters in the dataset represent the original values and the \textcolor{red}{red 'x'} characters denote the removed values. The horizontal red line implies that the entire row is removed (i.e., instance selection) and the vertical red line removes the column (i.e., feature selection).}
      \label{fig:fvs:probabilistic:flowchart}
      \bigVspace
\end{figure*}

The first algorithm, \methodA{}, is guided by the above defined removal probability. We visualize the complete process of \methodA{} in Figure \ref{fig:fvs:probabilistic:flowchart} to ease the understanding. We represent the original dataset using a table format, with columns corresponding to features, rows representing instances, and symbol ``v'' standing for a value (which could be missing). We then extract distinct values \xvalue{} for each feature $f^x$ to form the respective value set \XVALUE{} and then derive their probability $P(V^x \setminus v^x_z)$ corresponding to the given information metric $\iota$. Next, we use the probability to guide the value selection process. To be more specific, for each unique value \xvalue{} corresponding to a feature, we utilize a random generator to produce a score $r'$ between 0 and $\epsilon^{-1}$. If $r$ is smaller than the probability of \xvalue{}, value \xvalue{} will be removed from the respective value set \XVALUE{}. As shown in the third process in Figure~\ref{fig:fvs:probabilistic:flowchart} (refer this as ``Filtered Values''), the number of values in each feature might be reduced due to the value removal. Finally, we represent the dataset again using a table format, as shown in the last process of Figure~\ref{fig:fvs:probabilistic:flowchart}. Each symbol ``x'' indicates that the value originally located at this field has been removed.

Please note that the proposed method can mimic the ``feature selection'' or ``instance selection'' process,
%
%
as shown in the last step in Figure~\ref{fig:fvs:probabilistic:flowchart}. To generalize \fvs{} as an instance selection method, the number of missing value in an instance is evaluated. Note a missing value in an instance could be originally missing or removed via \fvs{}.
%
%
It is assumed that instances with more missing values are more irrelevant toward the training process, and thus, can be eliminated. Ultimately, an instance without any value is meaningless (i.e., all of the values are missing values) and can be removed without any reduction on the model's performance.
Similarly, \fvs{} can act as a feature selection when the values in a feature are completely removed, as the predictive power of that feature becomes null and thus, the feature can be safely removed. Finally, the time complexity of the \methodA{} method is bounded by $O(n+x)$, where $n$ and $x$ are the number of instances and the number of values in the dataset, respectively. 

\subsection{Extension of the Probabilistic \FVS}
\label{sec:fvs:method:probabilisticplus}

\methodB{} is an extension of \methodA{} by performing the \fvs{} per instance (locally), instead of applying the selection on all instances (globally). It is motivated by the following observation. Given a value \xvalue{} with non-zero information gain, it is expected to have a positive impact on the model's prediction power in some cases. If it is removed, we also lose its positive impact on those cases. That is to say if we select values at feature level, removing an actually useful value has an negative impact. In order to reduce these types of negative impact, we want to enable the value selection at the instance level but not the feature level. 
%
%

The complete process of \methodB{} is presented in Algorithm \ref{algo:fvs:probabilisticplus}. It scans through all the instances in the database. For each instance \pinstance{}, it checks all the non-missing values \xvalue{}s and decides whether to remove the value. The decision is purely dependent on the importance of this value (via either entropy or information gain metric) and a random score $r'$ generated by a random process (Line~\ref{algo-line:P+-random}). We simply compare the importance of this value with the generated random number, which is the same as what \methodA{} does. However, the real removal action is different. \methodB{} only removes the value from the current instance, and this removal will \emph{NOT} affect other instances (Lines~\ref{algo-line:P+-vs-start} - \ref{algo-line:P+-vs-end}). 

In addition to removing values at instance-level, \methodB{} also introduces the action of eliminating instances. For a given instance \pinstance{}, it derives the ratio of the number of features that instance \pinstance{} does not have values (either removed by the previous value selection process or originally missing) to the total number of features, namely \emph{missRate} in Line \ref{algo-line:P+-missrate}. Take instance $i_1$ in Table~\ref{tab:samples} as an example. Assume its values corresponding to both features $f^2$ and $f^4$ are removed; its value corresponding to feature $f^1$ is originally missing. Accordingly, its \emph{missRate} is $3/4=0.75$. \methodB{} generates a new random number $r'$ and removes the whole instance \pinstance{} if its \emph{missRate} is larger than the random number (Lines~\ref{algo-line:P+-random-2} - \ref{algo-line:P+-removeinstance}). When many values of an instance is missing, it is more likely that this instance has many high-uncertain/less-informative values. In other words, this instance has a higher chance to be noisy. We expect this instance-level removal could improve the capability of \fvs{} to remove unproductive instances and hence contributes to a more accurate classification performance. 
%

\begin{algorithm}[t]
\footnotesize
	\KwIn{Original instances: $I$, value set: $V$, amplifier $\epsilon$, information metric $\iota$}
	\KwOut{Filtered instances: $I'$}
	\BlankLine
    	Define $I' \leftarrow I$\;
        \ForEach{$i'_y$ in $I'$}{
            \ForEach{$v_z^x$ in $i'_y$}{
                 $r' \leftarrow random(0,1)$\; \label{algo-line:P+-random}
                \If{$\iota$=infoGain and $IG_N(D,v_z^x) < r' * \epsilon$} {\label{algo-line:P+-vs-start}
                    $i'_y$.remove($v$)\;
                }
                \ElseIf{$\iota$=entropy and $H(D|v_z^x) > r' * \epsilon$}{
                    $i'_y$.remove($v_z^x$)\;\label{algo-line:P+-vs-end}
                }
            }
            $missRate \leftarrow countMiss(i'_y)$\; \label{algo-line:P+-missrate}
            $r' \leftarrow random(0,1)$\; \label{algo-line:P+-random-2}
            \If{$missRate > r'$}{
                $I'$.remove($i'_y$)\; \label{algo-line:P+-removeinstance}
            }
        }
    return $I'$
\caption{\methodB{} method.}
\label{algo:fvs:probabilisticplus}
\end{algorithm}

\subsection{\FVS{} effectiveness proof}
\noindent
\textbf{Hypothesis}.
\Fvs{} will reduce, but will not increase, the confusion of the dataset by removing confusing values. 
%
%
We assume the proportion of each value \xvalue{} before and after the \fvs{} in feature \feature{} is represented by $w_z^x$ and $\widetilde{w}_z^x$, respectively, and the entropy of each respective value is described by $H(D|v^x_z)$.

\noindent
\textbf{Proposition}. Total confusion (i.e., weighted sum of the entropy) post-\fvs{} is no greater than that of the original data. 
\begin{equation}
\label{eq:fvs:proof}
    \sum_{x=1}^{|F|}\sum_{v=1}^{|V^x|} \widetilde{w}_z^x H(D|v^x_z) \leq \sum_{x=1}^{|F|}\sum_{v=1}^{|V^x|} w_z^x H(D|v^x_z)
\end{equation}
 
\begin{proof}

\begin{flalign*}
\widetilde{w}_z^x & = \left\{ \begin{array}{ll}
0 & \textrm{if } H(D|v^x_z) = 1 \\
w_z^x \times (1-H(D|v^x_z)) & \textrm{otherwise}
\end{array} \right.
\end{flalign*}
Since, $w_z^x \times (1-H(D|v^x_z))$ is always larger than 0,  we have 
\begin{flalign*}
\sum_{x=1}^{|F|}\sum_{v=1}^{|V^x|} \widetilde{w}_z^x H(D|v^x_z) &= \sum_{x=1}^{|F|}\sum_{v=1}^{|V^x|} w_z^x (1-H(D|v^x_z)) H(D|v^x_z) \\
&=\sum_{x=1}^{|F|}\sum_{v=1}^{|V^x|} w_z^x H(D|v^x_z) - \sum_{x=1}^{|F|}\sum_{v=1}^{|V^x|} w_z^x H^2(D|v^x_z)
\end{flalign*}
Accordingly, we can restate Inequality~(\ref{eq:fvs:proof}) as follows.
\begin{flalign*}
& \sum_{x=1}^{|F|}\sum_{v=1}^{|V^x|} w_z^x H(D|v^x_z) - \sum_{x=1}^{|F|}\sum_{v=1}^{|V^x|} \widetilde{w}_z^x H(D|v^x_z) \\
&= \sum_{x=1}^{|F|}\sum_{v=1}^{|V^x|} {w}_z^x H(D|v^x_z) - \sum_{x=1}^{|F|}\sum_{v=1}^{|V^x|} w_z^x H(D|v^x_z) +\sum_{x=1}^{|F|}\sum_{v=1}^{|V^x|} w_z^x H^2(D|v^x_z)) \\
&= \sum_{x=1}^{|F|}\sum_{v=1}^{|V^x|} w_z^x H^2(D|v^x_z) \ge 0
\end{flalign*}
Our proof completes. 
%
\end{proof}
\section{Experiments and Discussions}
\label{sec:fvs:exp}

In order to evaluate the performance of proposed approaches, we have conducted a comprehensive experimental study. In the following, we present the experiment setups, study the impact of hyper-parameters before we determine the exact setups, and then report our major findings. 
%

\subsection{Experiment Settings}
\label{sec:fvs:setting}

\begin{table*}[t]
\begin{center}
\footnotesize
\caption{The basic statistics of the \Ndataset{} benchmark datasets used in this paper.}
\label{tab:fvs:dataset}
\begin{tabular}{llrrrrrr}
\hline
ID & Dataset name & \#Instances & \#Features & \#Numerical & \#Categorical & \#Class & Missing values \\
\hline
d1 & credit & 1,000 & 20 & 7 & 13 & 2 & NO \\
d2 & hypothyroid & 3,772 & 29 & 7 & 22 & 4 & YES \\
d3 & mfeat-zernike & 2,000 & 47 & 47 & 0 & 10 & NO \\
d4 & segment-challenge & 1,500 & 19 & 19 & 0 & 7 & NO \\
d5 & letter & 20,000 & 16 & 16 & 0 & 26 & NO \\
d6 & adult & 48,855 & 14 & 6 & 8 & 2 & NO \\
d7 & census-income & 199,504 & 41 & 13 & 28 & 2 & NO \\
d8 & dota2 & 92,641 & 116 & 0 & 116 & 2 & YES \\
d9 & cifar10-small & 10,000 & 1,024 & 1,024 & 0 & 10 & NO \\
d10 & cifar10-big & 60,000 & 1,024 & 1,024 & 0 & 10 & NO \\
\hline
\end{tabular}
\end{center}
\bigVspace
\end{table*}


\noindent
\textbf{Datasets.} 
In our study, we use \Ndataset{} benchmark datasets from various domains \cite{Dua:2017, Krizhevsky09} with varying size in terms of the number of features, the number of instances, the possibility of missing values in the data, and the number of class labels. Table \ref{tab:fvs:dataset} reports the basic statistics of the datasets.

\noindent
\textbf{Algorithms.}
In order to evaluate the performance and effectiveness of proposed \methodA{} and \methodB{}, we implement in total \NBaseline{} representatives of \emph{feature selection (FS)}, \emph{feature transformation (FT)}, and \emph{instance selection (IS)} as competitors/baselines. They are i) \underline{FS\_CFS}~\cite{Hall1998}, a feature selection method that picks a set of useful features based on an evaluation formula with an appropriate correlation measure and a heuristic search strategy; ii) \underline{FS\_Consistency}~\cite{Liu1996}, a filter solution for feature selection using inconsistency metric and a probabilistic approach; iii) \underline{FS\_IWSS}~\cite{IWSS2009}, an incremental wrapper feature selection that first sorts features using a filter approach and evaluates them using a wrapper approach; iv)~\underline{FS\_MOEA}~\cite{MOEA2017}, a wrapper feature selection that employ genetic algorithm with two objectives: minimizing number of features and RMSE of Random Forests; v) \underline{FS\_PSO}~\cite{PSOSearch2007}, a feature selection that uses particle swarm optimization to find the optimal subset; vi)~\underline{FT\_RandomProjection} or \underline{FT\_RP} for short \cite{Bingham:2001:RPD:502512.502546}, a method that reduces the dimensionality of data by using a random projection from the original data space; vii) \underline{FT\_PCA}~\cite{pearson1901liii}, a data transformation method that projects the original dataset into a set of values of linearly uncorrelated label called principal components; viii) \underline{IS\_Misclassified}~\cite{hall2009weka}, an instance selection method implemented in Weka\footnote{\url{http://www.cs.waikato.ac.nz/ml/weka/}} that filters out instances that often mis-classify the class labels; and ix) \underline{IS\_Reservoir}~\cite{Vitter:1985:RSR:3147.3165}, a fast instance selection method that selects the instances using random sampling without replacement method (i.e., each instance in the dataset is stored into a set with a limited size using a probabilistic approach). Note the prefix (i.e., FS or FT or IS) indicates the category of the baseline


\noindent
\textbf{Parameters.} 
The parameter settings of different algorithms are explained as follows. The number of principal components in the \textit{FT\_PCA} is 1/2 of the number of features in the dataset and 95\% of variance is ensured to be in the original data. Similarly, the number of projected dimensions in the \textit{FT\_RP} is 1/2 of the number of features in the dataset. It is assumed that by reducing the number of features into a half of the original number of features, both the model size reduction and accuracy can be simultaneously high. The size of reservoir in \textit{IS\_Reservoir} is set to be 1/20 of the number of instances in the dataset. By sampling more instances from the dataset, the accuracy ratio is not necessarily higher than the case with fewer instances. Additionally, the original instances set in Weka consists of only 100 instances, which does not favor bigger datasets. Therefore, proportionally sampling 5\% of the dataset is more effective than fixing the size of the reservoir as it is difficult to determine a \textit{magic number} that works for all datasets. The other preprocessing methods are implemented using the default settings in Weka.

\noindent
\textbf{Setups.} 
All the experiments are conducted on a server with Intel Core i7-4790 running at 3.60GHz, 28 GB RAM, Windows 7, Java 9.0.1, and Weka 3.8.1.
%
%
\Fvs{} is implemented as a Weka's Filter class and Weka's classification methods are utilized to run the experiments. Our experiments can be divided into three stages namely \emph{discretization}, \emph{\fvs}, and \emph{classification}, as shown in Figure~\ref{fig:framework}. We evaluate three different discretization methods throughout our experiments to observe the robustness of the proposed methods under both unsupervised and supervised discretization methods, including \emph{equal-width binning (Binning)}, \emph{equal-frequency binning (Frequency)}, and \emph{minimum description length (MDL)} discretization~\cite{Fayyad1993}. For the sake of consistency, we implement all three discretization methods in Weka. Subsequently, we produce filtered data from the preprocessing of the discretized data. To cope with the randomness of produced results in \fvs, we repeat the experiments at random for five times. Finally, we construct the classification model using the filtered data and apply the 10-fold cross-validation scheme to evaluate the model.

\subsection{Parameter Analysis}

\label{sec:fvs:params}
\begin{table}[t]
\centering
\footnotesize
\caption{Hyperparameters}
\label{tab:hyper-parameters}
\begin{tabular}{l|c}
\hline
information metric $\iota$ & \textbf{entropy}, information gain \\ \hline
discretization method & Binning, \textbf{Frequency}, MDL \\ \hline
amplifier $\epsilon$ & 0.1, 0.2, 0.3, 0.4, \textbf{0.5}, 0.6, 0.7, 0.8, 0.9, 1.0
\\ \hline
\end{tabular}
\bigVspace
\end{table}

The performance of both \methodA{} and \methodB{} is dependent on several hyper-parameters, including the selected information metric $\iota$, the discretization method, and an amplifier $\epsilon$. In our last set of experiments, we study their impact on the performance of newly proposed algorithms. Table~\ref{tab:hyper-parameters} reports the settings of these three hyper-parameters, with values in bold indicating the default values. When we evaluate the impact of one hyper-parameter, we set the other two hyper-parameters to their defaults. Because of the space limitation, we only report their impacts on \methodB{}, as those hyper-parameters have similar impacts on \methodA{}. Also, borrowing the idea of F1 score that summarizes precision and recall, we define a \emph{harmonic mean} $\overline{X}$ of MR and AR as $\overline{X} = \frac{2}{\frac{1}{AR} + \frac{1}{MR}} = \frac{2 \times AR \times MR}{AR + MR}$. 

\begin{figure}[t]
      \centering
      \includegraphics[width=.9\linewidth]{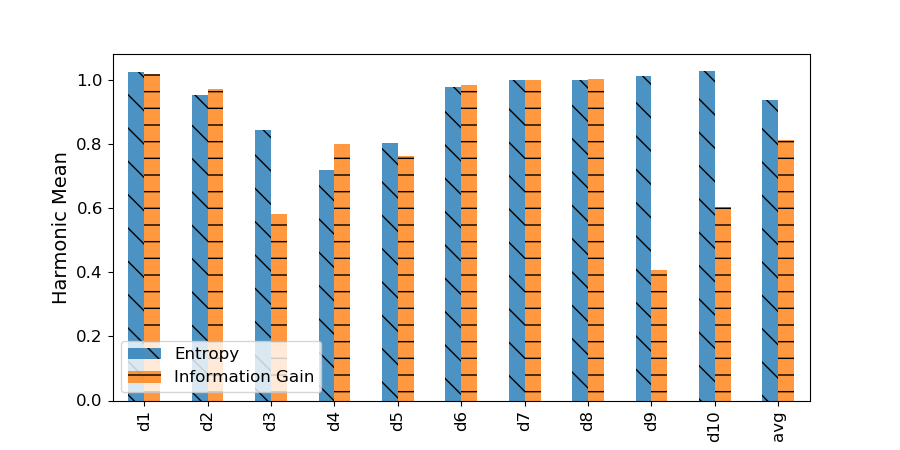}
      \caption{\Fvs{} (\methodB{}) performances under different information metric $\iota$ settings. Each point represents the performance on each dataset.}
      \label{fig:fvs:params:iota}
      \bigVspace
\end{figure}

\begin{figure}[t]
      \centering
      \includegraphics[width=.9\linewidth]{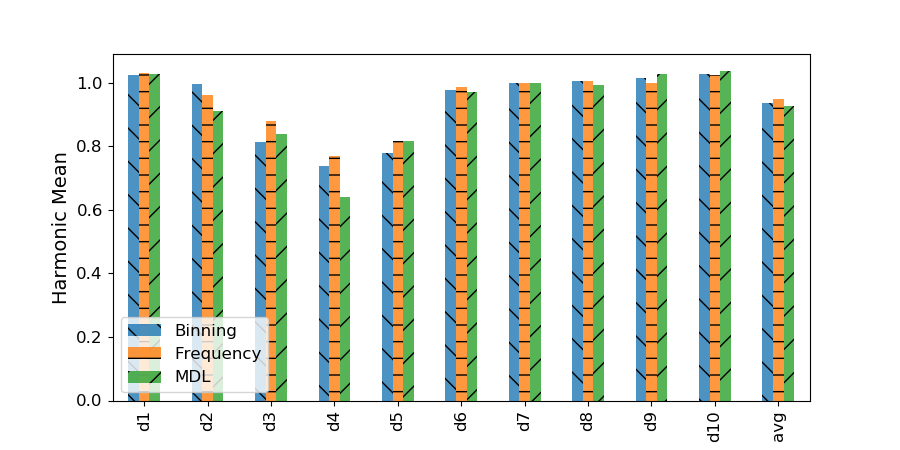}
      \caption{\Fvs{} (\methodB{}) performances under different discretization settings. Each point represents the performance on each dataset.}
      \label{fig:fvs:params:discretization}
      \bigVspace
\end{figure}

\begin{figure}[t]
      \centering
      \includegraphics[width=.9\linewidth]{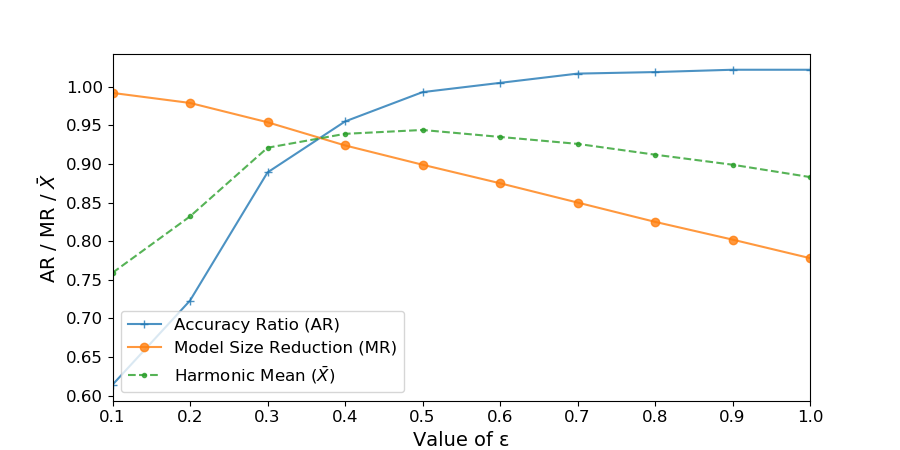}
      \caption{The impact of parameter $\epsilon$ on the \fvs{} methods (average performance on \Ndataset{} datasets).}
      \label{fig:fvs:params:epsilon}
      \bigVspace
\end{figure}

First, the impact of the information metric on \methodB{} is reported in Figure \ref{fig:fvs:params:iota}. While both entropy and information gain are commonly employed as measurements to evaluate the importance of data in the field of information theory, it seems that entropy is more useful to depict the usefulness of each value \xvalue{} than information gain in our study. 
%
Second, we evaluate the impact of discretization methods and report the results in Figure~\ref{fig:fvs:params:discretization}. There are no significant differences among three discretization methods, which demonstrates the robustness of the proposed \fvs{} methods on particular discretization methods.
Last but not least, we report the impact of amplifier $\epsilon$ in Figure~\ref{fig:fvs:params:epsilon}. As observed, it has opposite impacts on MR and AR. A small $\epsilon$ favors MR but not AR, while the increase of $\epsilon$ improves AR but not MR. The best trade-off between MR and AR is achieved when $\epsilon = 0.5$ (i.e., highest harmonic mean $\overline{X}$). 
%

\subsection{Overall Results}

\begin{figure*}[t]
      \centering
      \includegraphics[width=0.84\linewidth]{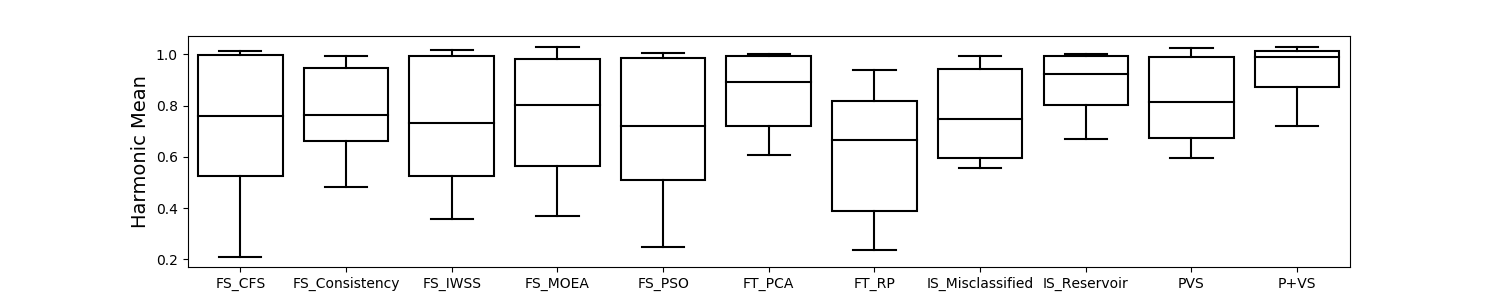}
      \caption{Harmonic mean of accuracy ratio and model size reduction of the proposed methods along with the baselines.}
      \label{fig:fvs:baselines}
\end{figure*}

\begin{table*}[t]
\centering
\footnotesize
\caption{Summary of Accuracy Ratio on Each Dataset}
\label{tab:fvs:accuracy_ratio}
\begin{tabular}{l|cccccccccc|c}
\hline
Algorithms & d1 & d2 & d3 & d4 & d5 & d6 & d7 & d8 & d9 & d10 & Average \\
 \hline
FS\_CFS & 1.07 & 0.99 & 1.02 & \underline{1.00} & 1.00 & 1.01 & \underline{1.01} & \underline{1.07} & 1.01 & 1.01 & 1.02 \\
FS\_Consistency & 1.07 & \underline{1.00} & 0.96 & \underline{1.00} & 0.99 & \underline{1.02} & \underline{1.01} & 1.02 & 1.03 & 1.03 & 1.01 \\
FS\_IWSS & \underline{1.10} & \underline{1.00} & \underline{1.03} & 0.99 & 1.00 & 1.01 & \underline{1.01} & \underline{1.07} & 1.03 & 1.03 & 1.03 \\
FS\_MOEA & 1.07 & \underline{1.00} & \underline{1.03} & \underline{1.00} & 1.00 & \underline{1.02} & \underline{1.01} & \underline{1.07} & 1.03 & N/A & 1.03 \\
FS\_PSO & 1.09 & \underline{1.00} & 1.01 & \underline{1.00} & \underline{1.01} & \underline{1.02} & \underline{1.01} & \underline{1.07} & 1.00 & 1.02 & 1.02 \\
FT\_PCA & 1.04 & 0.98 & 0.98 & 0.91 & 0.71 & 1.00 & 1.00 & 1.02 & N/A & N/A & 0.95 \\
FT\_RandomProjection & 0.98 & 0.98 & 0.42 & 0.72 & 0.39 & 0.96 & 0.99 & 0.98 & 0.77 & 0.70 & 0.79 \\
IS\_Misclassified & 1.07 & \underline{1.00} & 1.02 & \underline{1.00} & 0.99 & \underline{1.02} & \underline{1.01} & 1.04 & \underline{1.07} & 1.07 & 1.03 \\
IS\_Reservoir & 1.02 & 0.97 & 0.52 & 0.73 & 0.68 & 1.01 & 1.00 & 1.01 & 0.79 & 0.83 & 0.86 \\
\methodA{} & 1.06 & 0.97 & 0.96 & 0.94 & 0.84 & 0.95 & \underline{1.01} & 1.02 & 0.96 & 0.97 & 0.97 \\
\methodB{} & 1.06 & 0.96 & 0.90 & 0.92 & 0.91 & 0.97 & 1.00 & 1.00 & \underline{1.07} & \underline{1.08} & 0.99
\\ \hline
\end{tabular}
\end{table*}

\begin{table*}[t]
\centering
\footnotesize
\caption{Summary of Model Size Reduction on Each Dataset}
\label{tab:fvs:model_reduction}
\begin{tabular}{l|cccccccccc|c}
\hline
Algorithms & d1 & d2 & d3 & d4 & d5 & d6 & d7 & d8 & d9 & d10 & Average \\
 \hline
FS\_CFS & 0.96 & 0.81 & 0.4 & 0.46 & 0.34 & \underline{0.99} & \underline{0.99} & 0.96 & 0.12 & 0.14 & 0.62 \\
FS\_Consistency & 0.93 & 0.71 & 0.5 & 0.39 & 0.32 & 0.96 & 0.98 & 0.64 & 0.51 & 0.58 & 0.65 \\
FS\_IWSS & 0.94 & 0.75 & 0.42 & 0.43 & 0.34 & 0.95 & \underline{0.99} & 0.96 & 0.28 & 0.22 & 0.63 \\
FS\_MOEA & 0.89 & 0.7 & 0.48 & 0.39 & 0.32 & 0.95 & 0.97 & \underline{0.99} & 0.23 & N/A & 0.66 \\
FS\_PSO & 0.9 & 0.7 & 0.44 & 0.4 & 0.32 & 0.94 & 0.98 & 0.95 & 0.14 & 0.15 & 0.59 \\
FT\_PCA & 0.97 & 0.44 & 0.67 & 0.65 & 0.53 & 0.98 & 0.98 & \underline{0.99} & N/A & N/A & 0.78 \\
FT\_RandomProjection & 0.75 & 0.24 & 0.36 & 0.14 & 0.28 & 0.9 & 0.89 & 0.59 & 0.59 & 0.64 & 0.54 \\
IS\_Misclassified & 0.86 & 0.71 & 0.47 & 0.38 & 0.4 & 0.95 & 0.98 & 0.81 & 0.48 & 0.4 & 0.64 \\
IS\_Reservoir & 0.98 & \underline{0.96} & \underline{0.93} & \underline{0.85} & \underline{0.92} & \underline{0.99} & \underline{0.99} & 0.97 & 0.94 & 0.95 & 0.95 \\
\methodA{} & \underline{0.99} & 0.92 & 0.53 & 0.47 & 0.46 & \underline{0.99} & \underline{0.99} & 0.98 & 0.52 & 0.52 & 0.74 \\
\methodB{} & \underline{0.99} & 0.95 & 0.8 & 0.59 & 0.72 & \underline{0.99} & \underline{0.99} & \underline{0.99} & \underline{0.97} & \underline{0.99} & 0.90
\\ \hline
\end{tabular}
\end{table*}

\label{sec:fvs:results}

Our algorithms are designed to reduce the model sizes without hurting the accuracy achieved by the models. Consequently, we care both model size reduction (MR) and accuracy ratio (AR). The results in terms of $\overline{X}$ of all the algorithms are reported in Figure~\ref{fig:fvs:baselines}. Table \ref{tab:fvs:accuracy_ratio} and Table \ref{tab:fvs:model_reduction} report the detailed accuracy ratio and model size reduction of different algorithms on each dataset, respectively. To ease the understanding of the data, we underline the best performers corresponding to each dataset. Note that for both \methodA{} and \methodB{}, we report their performance with hyper-parameters set to their default (as listed in Table~\ref{tab:hyper-parameters}). 
%
%

%
%

We observe that our algorithms achieve good performance. To be more specific, \methodB{} achieves the best trade-off between AR and MR. If we take a closer look at its performance in different datasets, \methodB{} achieves the best AR performance in 2 out of 10 datasets. In the other 8 datasets, its AR performance is also comparable with the best performer, on average 6\% lower than the best performer. Its MR performance is even better. It tops in 6 out of 10 datasets. 

In terms of the comparison among \NBaseline{} competitors, we have several observations. Feature selection algorithms (i.e., FS\_CFS, FS\_Consistency, FS\_IWSS, FS\_MOEA, FS\_PSO) are designed to reduce the dimensionality of data and meanwhile to improve the model's performance. They indeed maintain impressive accuracy ratio. They are the best performer in terms of accuracy in 8 out of 10 datasets. However, they are not good in overall model size reduction, especially when the number of features is extremely large (e.g., d9, d10). On the other hand, IS\_Reservoir reduces the model size most but it suffers from unstable accuracy. It is worth noting that IS\_Reservoir drops its accuracy significantly in several datasets (e.g., d3, d4, d5, and d9). Meanwhile, we also observe that IS\_Misclassified, although being a representative of instance selection, performs more similarly to feature selection, because its priority is to remove mis-classified instances and hence focuses more on model's classification accuracy. FT\_PCA, as a feature transformation algorithm, tries to achieve a balance between accuracy ratio and model size reduction through the dataset transformation. Although it has not been the best performer in terms of AR or MR in any of the datasets, it does outperform many of the competitors in terms of the harmonic mean $\overline{X}$. Note we do not report the performance of FT\_PCA under datasets d9 and d10 (and FS\_MOEA on d10). This is because, under those two datasets, the memory usage of FT\_PCA (and FS\_MOEA) exceeds the capacity and its running time is extremely long. Finally, another feature transformation algorithm, FT\_RandomProjection, is faster than FT\_PCA and presents acceptable results in certain datasets but it fails to achieve overall good results because unlike FT\_PCA, it is not optimized and solely relies on randomization.
\section{Conclusions and Future Works}
\label{sec:fvs:conclusion}
A new alternative to the preprocessing methods in data mining is proposed by removing irrelevant values on the dataset. The purpose of the proposed method, \fvs, is reducing the model size and maintaining an acceptable accuracy ratio. Two probabilistic methods are presented to solve the \fvs{} problem: \methodA{} and \methodB{}. 
The former removes the values globally over all instances while the latter applies the removal locally on each instance. Moreover, \methodB{} can act as an instance selection by using the ratio of missing values in an instance.
Experiment results show that the proposed methods are effective in reducing the model size (59\%-99\% reduction) and simultaneously maintaining the accuracy ratio above 90\% on average. Furthermore, the proposed methods perform in a linear time complexity.

The direction of the future works is given as follows. 
Firstly, different heuristic metrics could be explored to generate the value subset.
In addition, combining \fvs{} with other data reduction methods could also be a good direction for inventing a better space efficient model. Tackling on more sophisticated problems, \fvs{} methods designed for online learning and semi-supervised learning are also an interesting direction.
\section*{Acknowledgements}
This research / project is supported by the National Research Foundation, Singapore under its International Research Centres in Singapore Funding Initiative. Any opinions, findings and conclusions or recommendations expressed in this material are those of the author(s) and do not reflect the views of National Research Foundation, Singapore.


\bibliographystyle{IEEEtran}
\bibliography{reference}

\end{document}